\pdfoutput=1
\documentclass[11pt]{article}
\usepackage{acl}
\usepackage{times}
\usepackage{latexsym}
\usepackage[T1]{fontenc}
\usepackage[utf8]{inputenc}

\usepackage{microtype}
\usepackage{svg}
\usepackage{booktabs}
\usepackage{multirow}
\usepackage{subcaption}
\newcommand{\smallerfont}{\fontsize{8.7}{11}\selectfont}

\title{Investigating grammatical abstraction in language models using \\few-shot learning of novel noun gender}

\author{Priyanka Sukumaran$^{1}$, Conor Houghton$^{1,*}$, Nina Kazanina$^{2,*}$\\
$^{1}$Faculty of Engineering, University of Bristol\\
$^{2}$Department of Basic Neurosciences, University of Geneva $^{*}$Equal contribution\\
\texttt{\{p.sukumaran,conor.houghton\}@bristol.ac.uk},\\
\texttt{nina.kazanina@unige.ch}}

\begin{document}
\maketitle
\begin{abstract}
Humans can learn a new word and infer its grammatical properties from very few examples. They have an abstract notion of linguistic properties like grammatical gender and agreement rules that can be applied to novel syntactic contexts and words. Drawing inspiration from psycholinguistics, we conduct a noun learning experiment to assess whether an LSTM and a decoder-only transformer can achieve human-like abstraction of grammatical gender in French. Language models were tasked with learning the gender of a novel noun embedding from a few examples in one grammatical agreement context and predicting agreement in another, unseen context. We find that both language models effectively generalise novel noun gender from one to two learning examples and apply the learnt gender across agreement contexts, albeit with a bias for the masculine gender category. Importantly, the few-shot updates were only applied to the embedding layers, demonstrating that models encode sufficient gender information within the word-embedding space. While the generalisation behaviour of models suggests that they represent grammatical gender as an abstract category, like humans, further work is needed to explore the details of how exactly this is implemented. For a comparative perspective with human behaviour, we conducted an analogous one-shot novel noun gender learning experiment, which revealed that native French speakers, like language models, also exhibited a masculine gender bias and are not excellent one-shot learners either. 
\end{abstract}

\section{Introduction}
Deep learning models of language have been shown to acquire non-trivial grammatical knowledge and match human levels of performance on natural language processing tasks. For example, LSTM \citep{Hochreiter1997LongMemory} and transformer models \citep{Vaswani2017AttentionNeed} trained on next-word prediction are able to parse complex syntactic structures that are thought to be essential to natural language \citep{Everaert2015StructuresSciences}. Language models have been shown to perform long-distance grammatical number \citep{Linzen2016AssessingDependencies,Goldberg2019AssessingAbilities} and gender agreement \citep{An2019RepresentationStudy,Lakretz2021MechanismsHumans}, even with intervening phrases \citep{Marvin2018TargetedModels,Mueller2020Cross-LinguisticModels,Hu2020AModels} and in grammatical but meaningless sentences \citep{Gulordava2018ColorlessHierarchically}. 

The human language ability is not limited to employing grammatical rules in familiar cases. Language acquisition studies have shown that humans are able to easily generalise and apply grammatical knowledge in relation to novel words from very few examples. For example, \citet{Berko1958TheMorphology} showed that young children can learn a non-word such as `wug\textsubscript{sg}' and easily infer its plural form `wugs\textsubscript{pl}' [wugz], and similarly `kich\textsubscript{sg}' to `kiches\textsubscript{pl}' [kichiz]. Numerous studies have also shown that children as young as three years old learn grammatical gender categories for new words using determiner-noun pairs \citep{Melancon2015RepresentationsChildren,Blom2006EffectsInflection}. Older children can spontaneously infer the appropriate morpho-syntactic feminine and masculine forms for French novel nouns in previously unencountered contexts \citep{Seigneuric2007TheFrench,Karmiloff-Smith1981AReference}. This demonstrates that humans have the ability to form linguistic abstractions that extend beyond having specific grammatical rules for individual words. 

To address whether small-scale LSTMs and transformer language models can achieve human-like grammatical abstractions, we design a word-learning experiment, inspired by psycholinguistics studies of language acquisition and generalisation. We introduce a novel noun into the embedding layer of our trained language model and investigate both few-shot learning abilities and the acquisition of abstract grammatical gender in French. Critically, the few-shot updates are isolated to the embedding layers. We assess the ability of language models to learn the gender category of a novel noun from a few examples in one grammatical agreement context, and then apply this knowledge in another agreement context during testing. This would indicate that models represent gender as an abstract category that is not tied to occurrences of specific syntactic contexts, and this information can be represented within the word-embedding space.

Across four experimental conditions, both models successfully acquired and generalised the gender of novel nouns after learning only one to two examples of gender agreement. Models effectively predicted noun-adjective and noun-participle agreement after encountering examples of the novel nouns with gender-marked articles `le\textsubscript{m}' or `la\textsubscript{f}'. However, we observed a gender bias: the gender prediction accuracy for feminine novel nouns remained consistently lower than the accuracy for masculine nouns, even after ten learning examples. The models also effectively generalised gender from noun-adjective and noun-participle agreement to a rarer context, noun-relative-pronoun agreement, exhibiting less gender bias and appropriately predicting `lequel\textsubscript{m}' or `laquelle\textsubscript{f}' agreement with nouns. 

Our findings suggest that (1) language models appear to represent grammatical gender as an abstract property, and (2) this information is encoded in the representation layers of language models and can be changed with few-shot updates. Further analysis into the patterns of weight change in the embedding layers during few-shot learning of gender revealed that both models primarily update the representation of the novel noun. Only the transformer, however, also updates the embeddings of the masculine determiner `le\textsubscript{m}' even when it was not present in the learning examples. This suggests that models may learn gender by updating related gender-marked words rather than assigning it as a core property of nouns like humans do.

Finally, for a comparative perspective with human behaviour on an analogous task, we conducted a one-shot novel noun gender learning experiment with 25 first-language French speakers. We show that humans also exhibited a masculine gender bias in a sentence completion task that required inferring the gender of novel nouns. While models and humans may rely on different mechanisms to abstract grammatical gender and perform syntactic generalisations, gender bias is evident in both. This commonality suggests that the bias could either be an inherent characteristic of French noun gender distribution or an efficient strategy for language and grammatical gender acquisition.

\section{Background}
Our study employs a word-learning paradigm to examine how language models generalise grammatical categories to novel nouns across syntactic contexts. We question whether they truly abstract grammatical properties beyond previous occurrences and specific syntactic contexts, or if they can only employ these features in familiar, repeated patterns of lexical units. Since we are interested in quantifying human-like generalisability in models, we focus on smaller models, training corpora, and vocabularies. Below, we briefly outline related work and discuss our choice of language models and gender agreement tasks used in our few-shot word-learning paradigm.

\subsection{Related work}
Studies investigating generalisation in pre-trained BERT models \citep{Devlin2019BERT:Understanding} have shown that they are able to generalise syntactic rules to low-frequency words as well as to new words acquired during fine-tuning. For example, \citet{Wei2021FrequencyTransformers} evaluated the effect of word frequency on subject-verb number agreement, and showed that BERT accurately predicts agreement for word pairs that do not occur during training. \citet{Thrush2020InvestigatingGeneralization} showed that pre-trained BERT models are able to learn new nouns and verbs from a few learning examples and generalise linguistic properties related to both syntax and semantics in two aspects of English verbs: verb/object selectional preferences and verb alternations \citep{Levin1993EnglishAlternations}. 

\citet{Wilcox2020StructuralModels} investigated similar syntactic generalisations in RNN models; they showed that RNNs with structural supervision and unsupervised LSTMs can predict subject-verb agreement for low-frequency nouns appearing as few as two times in the training corpus. While models successfully generalised number agreement rules to low-frequency nouns, they exhibited a bias for transitive verbs, which was also seen in the BERT study \citep{Thrush2020InvestigatingGeneralization}.

To our knowledge, only one other study has focused on the generalisation and representation of grammatical categories in language models, and how this is extended to novel words. \citet{Kim2021TestingModels} investigated this in BERT models, and showed that they can infer the grammatical category of novel words from linguistic input that unambiguously categorises the novel word into noun, verb, adjective and adverb categories. However, they found that BERT required up to 50 fine-tuning iterations with a high learning rate to distinguish these categories during testing.

Our study adds to the current literature in three ways. Firstly, we assess syntactic rule generalisation using grammatical gender: a largely semantically arbitrary, inherently lexical property which is consequential in various grammatical agreement contexts in French. To our knowledge, grammatical gender agreement has not been previously tested in a novel-noun learning paradigm. Secondly, while previous studies have used either fine-tuning methods or analysed syntactic agreement of low-frequency words, we introduce novel word embeddings and isolate few-shot learning to the representation layers of language models, as done in \citet{Kim2021TestingModels}. This is more in line with the psycholinguistic hypotheses for linguistic generalisation in humans, whereby a set of grammatical agreement rules and categories are learnt, and new words are integrated with minimal changes into the broader linguistic knowledge. Third, we choose to train a smaller-scale, unidirectional LSTM and decoder-only transformer language model using training corpora that are better aligned with human language exposure. This provides a fairer comparison of model to human generalisation behaviour. 

\begin{table*}
\smallerfont
\centering
\begin{tabular}{@{}lll@{}}
\toprule
 & \textbf{Learning example} & \textbf{Test} (0-1 words between noun and target) \\ \midrule
 \textbf{A} & article-noun & noun-adjective \\
 & \begin{tabular}[c]{@{}l@{}}j'ai vu \textbf{le}\textsubscript{m}/\textbf{la}\textsubscript{f} \underline{\textbf{noun}} (I saw \textbf{the}\textsubscript{m/f} \underline{\textbf{noun}}) \end{tabular} & 
 \begin{tabular}[c]{@{}l@{}}je ne vois pas de \underline{\textbf{noun}} \textit{\textbf{vert}\textsubscript{m}/\textbf{verte}\textsubscript{f}} (I don't see a \textbf{green}\textsubscript{m/f} \underline{\textbf{noun}})\end{tabular} \\ \midrule
\textbf{B} & article-noun & noun-participle \\
 & \begin{tabular}[c]{@{}l@{}}j'ai vu \textbf{le}\textsubscript{m}/\textbf{la}\textsubscript{f} \underline{\textbf{noun}} (I saw \textbf{the}\textsubscript{m/f} \underline{\textbf{noun}}) \end{tabular} & 
 \begin{tabular}[c]{@{}l@{}}je ne vois pas de \underline{\textbf{noun}} \textit{\textbf{fixé}\textsubscript{m}/\textbf{fixée}\textsubscript{f}} (I don't see a \textbf{fixed}\textsubscript{m/f} \underline{\textbf{noun}})\end{tabular} \\ \midrule
\textbf{C} & noun-adjective & noun-relative-pronoun \\
 & \begin{tabular}[c]{@{}l@{}}je vois l'\underline{\textbf{noun}} \textbf{noir}\textsubscript{m}/\textbf{noire}\textsubscript{f} (I see the \textbf{black}\textsubscript{m/f} \underline{\textbf{noun}}) \end{tabular} & 
 \begin{tabular}[c]{@{}l@{}} je vois l'\underline{\textbf{noun}} sur \textit{\textbf{lequel}\textsubscript{m}/\textbf{laquelle}\textsubscript{f}}  
 (I see the \underline{\textbf{noun}} on \textbf{which}\textsubscript{m/f})\end{tabular} \\ \midrule
\textbf{D} & noun-participle & noun-relative-pronoun \\
 & \begin{tabular}[c]{@{}l@{}}je vois l'\underline{\textbf{noun}} \textbf{brisé}\textsubscript{m}/\textbf{brisée}\textsubscript{f} (I see the \textbf{broken}\textsubscript{m/f} \underline{\textbf{noun}}) \end{tabular} & 
 \begin{tabular}[c]{@{}l@{}} je vois l'\underline{\textbf{noun}} sur \textit{\textbf{lequel}\textsubscript{m}/\textbf{laquelle}\textsubscript{f}} (I see the \underline{\textbf{noun}} on \textbf{which}\textsubscript{m/f})\end{tabular} \\ \bottomrule
\end{tabular}
\caption{Example sentence constructions for few-shot learning and testing}
\label{tab:tests}
\end{table*}

\subsection{Grammatical agreement}
Grammatical agreement is a feature of many languages. In grammatical agreement, the
properties of nouns, such as number (singular/plural), determine and modify the form of other
words in the sentence, such as the verb, determiner or adjective. In morphologically rich
languages, agreement rules extend to other properties like gender, animacy, case or person. Psycholinguistic studies have used agreement tasks to probe the human ability to parse hierarchical syntactic structures in language \citep{Franck2002Subject-verbHierarchy}. This is because grammatical agreement relies on syntactic structure and cannot be deduced from linear word order in a sentence or word co-occurrence statistics \citep{Everaert2015StructuresSciences}. Consider the following short sentence in English and French, where the main noun `table' dictates the number (sg: singular, pl: plural) in both languages, and the gender (f: feminine, m: masculine) in French:
\begin{quote}
\small
La\textsubscript{sg.f} \textbf{table}\textsubscript{sg.f} [près des lits\textsubscript{pl.m}] est\textsubscript{sg} verte\textsubscript{sg.f} \\
(The \textbf{table}\textsubscript{sg} [near the beds\textsubscript{pl}] is\textsubscript{sg} green) 
\end{quote}
The above example shows how the noun `beds'/`lits' directly precedes the verb and adjective but does not trigger grammatical agreement, highlighting the importance of structure and syntactic properties over linear sequence for agreement. 

Grammatical agreement, in general, tests syntactic parsing and abstraction of agreement rules beyond specific examples encountered in the training corpus. Language models have been extensively evaluated on grammatical number agreement tasks \citep{Gulordava2018ColorlessHierarchically, Linzen2016AssessingDependencies}, see \citet{Linzen2021SyntacticLearning} for a comprehensive review; it has been shown that models can establish agreement even in complex and long-distance constructions. 

We propose that grammatical gender agreement additionally offers a more direct probe of linguistic abstraction. Differing from number, which is a semantically interpretable property that has a meaning in the real world, singular referring to one and plural referring to more than one, grammatical gender is often a semantically non-interpretable and idiosyncratic property of the noun \citep{Audring2014GenderFeature,Acuna-Farina2009TheOverview}, especially in French. Grammatical gender is thus a more abstract category than number, but only a few language modelling studies have focussed on it  \citep{An2019RepresentationStudy,Lakretz2021MechanismsHumans,Perez-Mayos2021AssessingModels}. 

Humans form an abstract representation of gender; do models also form it? In order to test the ability of models to perform grammatical agreement during sentence generation, we use the targeted syntactic evaluation approach \citep{Linzen2016AssessingDependencies,Futrell2019NeuralState}. Specifically, we assess model behaviour on test sentences that are carefully constructed to probe grammatical gender agreement. For example, given a test sentence requiring noun-adjective agreement like `La\textsubscript{sg.f} \textbf{table}\textsubscript{sg.f} est\textsubscript{sg}...', if the model assigns a higher probability to the correct adjective `verte\textsubscript{sg.f}' that agrees in number and gender with the head noun, compared to the grammatically incorrect alternative `vert\textsubscript{sg.m}', we consider this as successful use of grammatical properties for agreement.

\subsection{Language Models}
Pre-trained transformer models like BERT and GPT-3 \citep{Devlin2019BERT:Understanding,Brown2020LanguageLearners,Alec2019LanguageLearners} excel at various linguistic tasks \citep{Hu2020AModels} largely due to their ability to scale to billions of parameters and handle extensive data, often exceeding human language exposure. On the other hand, LSTMs often have far fewer parameters and mirror aspects of human language processing \citep{Hochreiter1997LongMemory,Elman1990FindingTime}. LSTMs operate on a sequential basis, mimicking constraints observed in human working memory processes and learn efficiently from limited corpora \citep{Ezen-Can2020ACorpus}. Our study will focus on uni-directional models that use incremental processing of language \citep{Christiansen2015TheLanguage,Cornish2017SequenceLearning}, which are more conducive to examining human-like language processing and generalisation. Specifically, we use LSTMs and a smaller-scale, decoder-only transformer model.

\section{Method}
\subsection{Model architectures and training}
We trained an LSTM and a decoder-only transformer language model with a next-word prediction objective in French. The LSTM, as described in \citet{Gulordava2018ColorlessHierarchically}, consisted of two hidden layers of 650 units each, and a vocabulary size of 42,908. LSTMs with similar specifications have been shown to predict noun-adjective and noun-participle agreement in French \citep{An2019RepresentationStudy,Sukumaran2022DoRules} and Italian \citep{Lakretz2021MechanismsHumans} even with attractor phrases. 

For the transformer, we trained a decoder-only architecture similar to GPT-1 \cite{Radford2018ImprovingPre-Training, Vaswani2017AttentionNeed}, with masked self-attention heads and positional encoding. The model had 12 layers, 12 heads, an embedding and hidden size of 768, and was trained over 50 epochs using SGD with a warm-up epoch followed by cosine learning rate annealing (See Appendix \ref{sec:appendixB} for details). While both SGD and AdamW achieved similar perplexities (supplementary Table \ref{tab:model_params}), training with SGD outperformed on the gender agreement baseline (Section \ref{sec:knownnouns}). This training approach aligns with \citet{Li2023AssessingAgreement}.

Although our transformer model has a much larger parameter space than our LSTM model, both models were trained using word-based tokenisation on identical corpora and vocabulary sizes for better comparability of model performance. The training corpora contained 80 million word tokens for training and 10 million tokens each for validation and testing, extracted from French Wikipedia sources \citep{Mueller2020Cross-LinguisticModels}, Appendix \ref{sec:appendixA}. This approximates human exposure during language acquisition; according to \citet{Gilkerson2017MappingAnalysis}, children encounter up to 7 million words each year. If we consider that major language acquisition takes place up to adolescence (age 10-12), the dataset would contain 70-84 million words \citep{Warstadt2023CallCorpus}. We also tied the weights between the input/output and embedding layers in both models. These layers perform analogous operations: mapping from one-hot encoded token vectors to dense embeddings and vice versa \citep{Press2017UsingModels}. As our experiment is aimed at evaluating the role of the representational layer in encoding grammatical gender information, weight tying may provide a more interpretable result where the word embeddings are the same between input and output. All results presented below are averages across three model instantiations.

\subsection{Novel nouns}
To test the ability of the models to learn the gender of previously unseen nouns, we create novel noun embeddings by combining the embeddings of two semantically similar existing nouns with opposite genders. This combination is performed by averaging vectors in the embedding space where $\textbf{x}($noun$)$ represents a vector that embeds a noun. For example, we can combine noun$_1=$`bague\textsubscript{f}' (ring) which is feminine and noun$_2=$`bracelet\textsubscript{m}' (bracelet) which is masculine: 
 \begin{equation}
\mathbf{x}(\rm{noun}) \leftarrow 0.5\mathbf{x}(\rm{noun}_1) +0.5\mathbf{x}(\rm{noun}_2).
\end{equation}
We insert $\textbf{x}($noun$)$ in place of the embedding of the least common token in the vocabulary to test it with minimal interference to the trained model. Prior to any learning steps, we assess the initial gender of the novel noun by evaluating gender prediction on test phrases such as `je ne vois pas de \textbf{\underline{noun}} \textit{\textbf{vert}\textsubscript{m}}/\textit{\textbf{verte}\textsubscript{f}}'. The gender of a novel noun is categorised as initially feminine if the LSTM assigns higher probability to the feminine target-word, e.g. `verte\textsubscript{f}' (green\textsubscript{f}) than its masculine alternative `vert\textsubscript{m}' (green\textsubscript{m}) and vice versa. We created a set of ten initially feminine and ten initially masculine novel nouns (Appendix \ref{sec:appendixC}). 

\subsection{Few-shot learning and testing}
Few-shot learning was implemented as a single gradient update with a training mini-batch of one to ten learning sentences. Crucially, the gradient is only applied to the embedding layers of the trained language model while the hidden layers and other components of the LSTM or transformer were kept unchanged. Thus, the language model was tasked with learning and generalising the novel noun's gender without making any modifications to the trained model structure.

The learning sentences contained the novel noun and set its gender using one of several grammatical constructions: article-noun (Conditions \textbf{A} and \textbf{B}), noun-adjective (Condition \textbf{C}) and noun-participle in (Condition \textbf{D}), see Table~\ref{tab:tests} for examples. For Conditions \textbf{C} and \textbf{D}, the gender information was provided by the adjective or participle; to avoid providing an extra gender cue using a gendered article, the gender-neutral article ‘l'' was used with the novel noun; `l'' is a contraction of both ‘le\textsubscript{m}’ and ‘la\textsubscript{f}' used with nouns starting with a vowel and thus does not reveal gender. This approach allowed for learning sentences in Condition \textbf{C} like ‘je vois l'\underline{\textbf{noun}} \textbf{noir}\textsubscript{m}/\textbf{noire}\textsubscript{f}', where the gender of a vowel-initial noun is revealed solely by the adjective's form. Few-shot learning was implemented with mini-batches of 1, 2, 3, 5, and 10 examples of a given learning construction. Each set was repeated five times with a new mini-batch of randomly selected subsets of examples from a total of 15. See Appendix \ref{sec:appendixD} for all the learning examples.

\begin{figure*}[h]
  \centering
  \includegraphics[width = \linewidth]{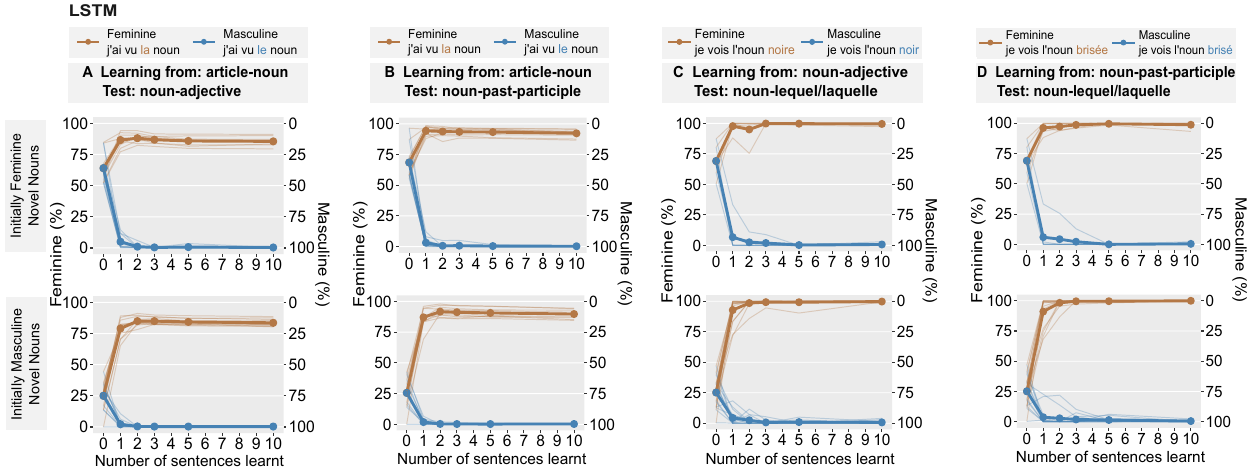}
  \caption{LSTM performance on gender agreement tests before learning, corresponding to zero sentences learnt, and after few-shot learning with 1, 2, 3, 5 and 10 sentences. The dark orange lines plot average prediction accuracy after learning from feminine training sentences, while the blue lines correspond to learning masculine sentences. The faded lines indicate the individual performance of 20 novel nouns. The left $y$-axis shows the prediction accuracy for feminine gender, while the right $y$-axis displays masculine gender accuracy such that $100\%$ accuracy for feminine gender corresponds to $0\%$ for masculine gender. Error bars of 95\% bootstrapped confidence intervals are too small to be seen.}
\label{fig:lstm_learning}
\end{figure*}

In each condition, the novel noun's gender was tested in a different gender agreement context from the one used in the learning construction. In learning Conditions \textbf{A} and \textbf{B}, the gender of the novel noun is inferred from article-noun agreement (indicated by `le\textsubscript{m}' or `la\textsubscript{f}') and tested using noun-adjective (\textbf{A}) or noun-participle agreement (\textbf{B}). In Conditions \textbf{C} and \textbf{D}, the learning construction used noun-adjective (\textbf{C}) and noun-participle (\textbf{D}) agreement, and the test construction involved sentences where the noun gender agrees with a relative pronoun: `lequel\textsubscript{m}' or `laquelle\textsubscript{f}'. In addition, to test adjacent vs non-adjacent or long-distance agreement, we varied the number of intervening words between the noun and target in each condition with 0-6 gender-neutral words. Accuracy scores in Figures \ref{fig:lstm_learning} and \ref{fig:transformer_learning} are based on the average across 120 test sentences; details are provided in Appendix \ref{sec:appendixE}.

\begin{figure}[h]
  \centering
  \includegraphics[width = \linewidth]{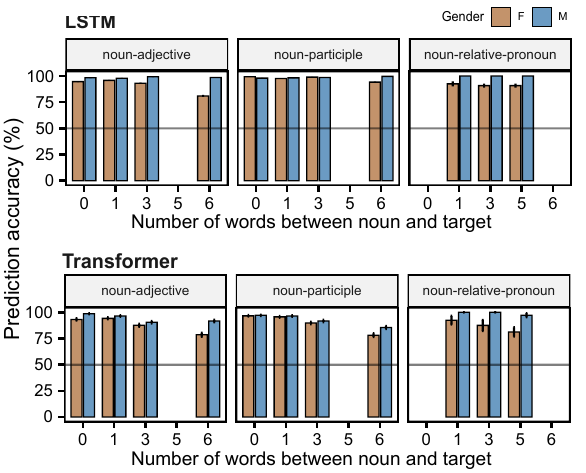}
  \caption{LSTM (top) and transformer (bottom) prediction accuracies of gender agreement with existing French nouns that appear in training data, across three agreement tests. Error bars are 95\% confidence intervals across sentences.}
  \label{fig:baseline}
\end{figure}

\section{Results}
\subsection{Baseline performance with known nouns}
\label{sec:knownnouns}
To ensure that both models can perform the baseline task of grammatical gender agreement, we tested gender prediction on existing 20 masculine and 20 feminine nouns that appeared at least 50 times in the training corpus. Both models consistently predicted gender agreement with accuracies well above chance (50\%) across three agreement constructions: \textbf{A} noun-adjective, \textbf{B} noun-participle and \textbf{C} noun-relative-pronoun agreement, see Figure \ref{fig:baseline}. The transformer model showed slightly lower average performance, $91.6\%\pm0.005$, than the LSTM, $96.4\%\pm0.001$.  Accuracy of predicting feminine gender agreement was $4.41\%$ lower than masculine gender for the LSTM, and $4.24\%$ for the transformer. While the LSTM effectively maintains long-distance agreement even with six intervening words between noun and target, the transformer's performance gradually declines by more than 10\% when the number of intervening words increases from zero to six. However, in Condition \textbf{A} with six intervening words, the LSTM exhibits a large gender bias of $7.36\%$, possibly indicating difficulty with this sentence construction involving a temporal modifier and relative clause.

\subsection{Few-shot learning of novel nouns}
\begin{figure*}[h]
  \centering
  \includegraphics[width = \linewidth]{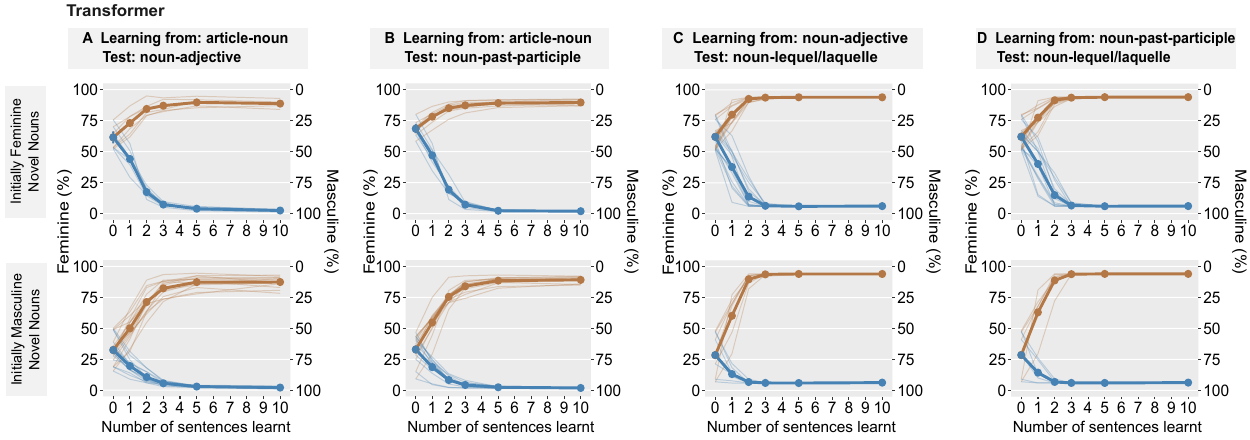}
  \caption{Transformer performance on gender agreement tests before and after few-shot learning. See Figure \ref{fig:lstm_learning} caption for details on layout, axes and content of graphs.}
\label{fig:transformer_learning}
\end{figure*}

Next, we test the language models on few-shot learning with 1, 2, 3, 5, and 10 examples from Conditions \textbf{A} and \textbf{B}. After learning from a single example signifying the masculine gender of a novel noun (Figure \ref{fig:lstm_learning}), average prediction accuracy rose to $96.5\%\pm0.01$ in Condition \textbf{A} and $97.5\%\pm0.01$ in Condition \textbf{B} for both initially feminine and masculine nouns. Learning feminine gender proved less efficient, yielding $82.4\%\pm0.02$ and $90.6\%\pm0.01$ accuracy in Conditions \textbf{A} and \textbf{B} respectively. Transformer performance (Figure \ref{fig:transformer_learning}) displayed a similar gender bias, with gradually increasing accuracy from one to five learning examples reaching $88.6\%\pm0.002$ for feminine and $98.0\%\pm0.001$ for masculine gender categorisation. This slower learning trajectory in the transformer is due to the choice of learning rate used during few-shot updates; see supplementary Figure \ref{fig:learningrate}. Beyond ten learning examples, accuracy improvement for both models is marginal.

In Condition \textbf{C}, after only one training example, the LSTM achieves a prediction accuracy of $94.3\%\pm0.001$ for feminine and $94.9\%\pm0.001$ for masculine learning trials. In Condition \textbf{D}, the accuracies are $95.6\%\pm0.001$ and $92.5\%\pm0.002$, respectively. With five to ten learning examples, the model's accuracy reaches up to $99\%$ in both feminine and masculine learning trials. The transformer model had a similar pattern of results with an average accuracy of $93.9\%\pm0.003$ after 5 learning examples in Condition (\textbf{C}), $94.0\%\pm0.002$ in Condition (\textbf{D}). Importantly, a learning bias with gender category was not seen.

\begin{figure}[h]
  \centering
  \includegraphics[width = \linewidth]{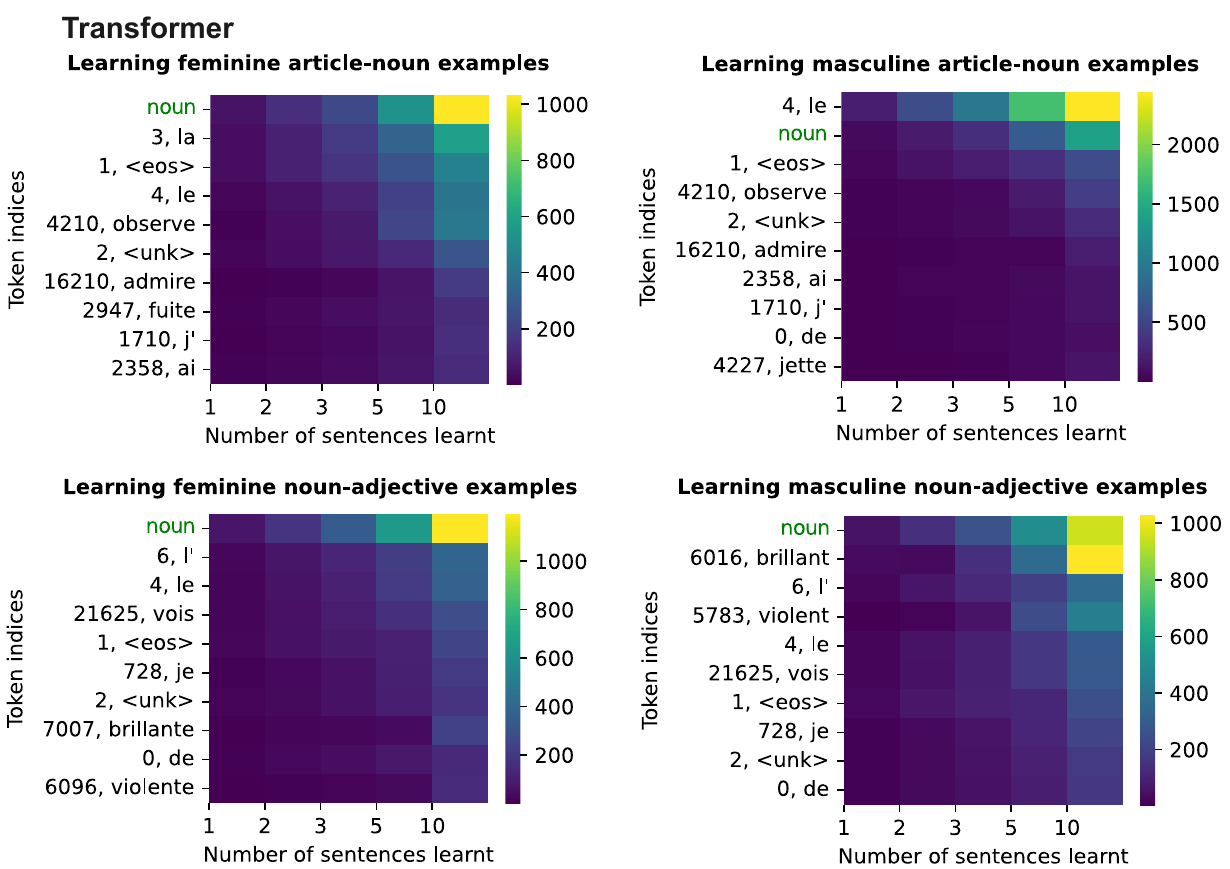}
  \caption{Transformer: Top ten tokens by percentage of weight change to embedding layer after few-shot learning updates with 1-10 sentences. Note that the colour scales representing percentage change are different in each panel.}
\label{fig:transformer_dw}
\end{figure}

\subsection{Weight changes to the novel noun}
To better understand the mechanisms underlying few-shot learning and grammatical generalisation, we analysed tokens with substantial weight changes to the embedding layer during learning. See Figure \ref{fig:transformer_dw} for weight changes in Conditions \textbf{A} and \textbf{C} for the transformer. Results for Condition \textbf{D} is included in supplementary Figure \ref{fig:transformer_dw_D}, showing similar patterns to \textbf{C}. The top ten tokens by magnitude of weight change included the novel noun and words present in the learning sentences. Notably, the transformer model also significantly adjusted the weight for the masculine article `le\textsubscript{m}', even if it was not in the learning sentences, and with the feminine learning condition. This is not due to the frequency of `le\textsubscript{m}' being 4\textsuperscript{th} in the vocabulary, since the feminine article `la\textsubscript{f}', 3\textsuperscript{rd} in the vocabulary, did not incur strong updates. In contrast, the LSTM's weight changes corresponded closely with tokens in learning examples; see supplementary Figure \ref{fig:lstm_dw}.

In a related analysis, for a given novel noun, we measured the weight change along the gender direction. We define the gender direction as the vector difference of two original noun embeddings that were used in the composition of the novel noun. We confirm that projections of the nouns' weight change along the gender axis were consistently negative (feminine), positive (masculine), or near zero in a gender-neutral control case according to the learning examples. 

\section{Human behavioural experiment} 
Although language models are vastly different from humans with regards to inductive biases and working memory constraints, comparing their performance and mechanisms to humans is useful for exploring possible strategies for grammatical representation in each system, see \citet{Saxe2021IfQuestion} for a review. In order to compare model performance on the generalisation task to human behaviour, we conducted a one-shot word learning experiment online, with 25 native French speakers. Participants learnt 16 novel nouns with a gender-ambiguous endings and 8 novel nouns with typical gender endings, shown in Table \ref{tab:human_exp_nouns}, adopted from \citet{Seigneuric2007TheFrench}. The nouns were presented in learning construction as in Table \ref{tab:tests} and participants were asked to complete a test sentence which required grammatical gender agreement. The learning example remained visible on the screen to alleviate memory load, see supplementary Figure \ref{fig:human_exp_slides}. We chose a sentence completion task to ensure that participants do not intently pay attention to the gender clues, which would become too trivial. We endeavoured to test all four conditions given the constraints of designing an analogous experiment for humans. The experiments revealed that humans achieve near perfect scores when predicting gender for existing French nouns, but fall short compared to models at one-shot learning of novel noun gender. While average scores were still above $75\%$ for equivalent Conditions \textbf{A} and \textbf{B}, humans participants exhibited a strong masculine bias when completing noun-relative-pronoun agreement in Conditions \textbf{C} and \textbf{D}, with feminine agreement accuracy almost at chance level. See Figure \ref{fig:human_exp} for results and Appendix \ref{sec:appendixH} for more details of the experiment.

\section{Discussion}
The primary goal of this work is to investigate whether language models develop an abstract grammatical gender category. To address this, we demonstrated that LSTMs and transformers are proficient in acquiring the gender of a novel noun from one to two learning examples, and apply gender agreement in a previously unencountered context. 

\paragraph{Both language models seem to acquire abstract gender properties of novel nouns from few-shot learning.} While our transformer exhibited marginally lower accuracies in the baseline gender agreement tasks, its few-shot learning capabilities and patterns are similar to the LSTM. More specifically, few-shot updates to the embedding layers are enough for acquiring novel noun gender and generalising this to unseen agreement contexts. This aligns with how humans are believed to learn words, which only requires an incremental update to the knowledge of nouns during acquisition while maintaining an abstract understanding of grammatical gender and agreement rules. It appears that language models have a similar capacity to generalise grammatical gender to include new words and that important grammatical category information may be encoded in the word embedding space learned by models. This is consistent with \citet{Kim2021TestingModels}, who demonstrate that noun, adjective, adverb and verb categories emerge in the model's representational space, and \citet{Lakretz2021MechanismsHumans}, who showed using principle component analysis that noun, adjective, verb and article embeddings encode gender and number properties.

Although models seem to succeed on tasks that require having a representation of grammatical gender that generalises across syntactic agreement contexts and extends to novel nouns, the specific implementation details are not immediately clear. We conducted additional analysis on weight changes and learning dynamics as an initial step to understand the underlying mechanisms of how gender is represented and generalised. For both models, few-shot learning primarily results in updates to embeddings of the novel noun and words that appear in the learning constructions. The transformer model additionally updates the representations of the gender-marked article, specifically `le\textsubscript{m}'. This suggests that the trained language model may not represent gender as critically hosted by nouns, and gender agreement as triggered by nouns alone. It may be that the transformer has developed a representational space governed by co-occurrence patterns, consisting of the word-embeddings, that groups nouns, verbs, adjectives and determiners such as `le\textsubscript{m}'/`la\textsubscript{f}' by gender. On the other hand, humans typically assign gender to nouns in a deterministic manner; where agreement is determined by the noun's gender and relies less on heuristics; this does not seem to be the case for our models, especially the transformer. An interesting parallel can be drawn between this mechanism in transformers and a child's acquisition of gender. Driven by their affinity to learn chunks of words, children begin to acquire noun gender through determiner-noun pairs, treating them as single units \citep{Mills1986TheGerman.,MacWhinney1978TheMorphophonology}; transformers seem to employ a similar strategy to encode word co-occurrence patterns. 

However, our weight change analysis alone does not provide conclusive evidence for exactly \textit{how} the model represents gender within its embedding layer, and whether it is truly abstract. Future work may investigate this by conducting additional few-shot learning experiments with weight updates restricted to the novel noun embeddings and other parts of the embedding space; this would reveal whether abstract grammatical gender is localised to a sub-space or to a single noun embedding in the representational space. Similarly, running the experiments with the embedding layer frozen while updating the rest of the model and comparing this with updating the whole model could reveal how important the representational layers are for grammatical gender and agreement. 

\paragraph{Further mechanistic explorations are required to understand the extent to which models form abstract grammatical gender.} For example, \citet{Lakretz2019TheModels} used causal mediation analysis to uncover sparse mechanisms whereby individual units in the LSTM tracked grammatical number and gender \citep{Lakretz2021MechanismsHumans}. \citet{Vig2020InvestigatingAnalysis} used similar methods to isolate gender bias to a group of attention heads in transformers. Future work could utilize similar methods to characterise how gender information from word embeddings is processed through the model to drive downstream agreement performance; this could reveal how influential and abstract the representation of grammatical properties is.

\paragraph{Language models exhibit masculine gender bias across four gender agreement contexts and during few-shot learning of novel noun gender.} On the baseline gender agreement task, transformers and LSTMs, to a lesser extent, exhibited a masculine gender bias. The bias could not have been due to frequency as the 20 feminine and 20 masculine nouns had similar frequencies in the corpus. Few-shot learning behaviour also showed this bias, where feminine gender prediction falls short of masculine prediction accuracies even after training with ten learning examples. One explanation for gender bias might be that it is an inherent property of French or the corpus. It may be because there are more masculine words \citep{Ayoun2018GrammaticalMyth}. Moreover, in colloquial French, past-participles and adjectives are produced in their default singular-masculine forms, omitting the plural/feminine inflections \citep{Belletti2007PastAgreement}, thus not obeying the agreement rule; it is likely that the corpus reflects this pattern. The observation of gender bias in language models is consistent with studies by \citet{Marvin2018TargetedModels} and \citet{Jumelet2019AnalysingAssignment} demonstrating that models encode a preferential or `default' category for grammatical properties: default singular number category and default masculine gender category. 

\paragraph{Humans are not perfect one-shot learners of novel noun gender either.} Given the numerous studies demonstrating acquisition of grammatical gender in 3-4 year old children \citep{Walter2021GrammaticalNouns,Seigneuric2007TheFrench,Eichler2013GenderItalian-French}, it is surprising that adult French speakers did not achieve high accuracies in inferring novel noun genders in our experiment. They also exhibited a masculine gender bias, like the language models. It is important to consider the experimental constraints that make it difficult to observe people's true generalisation abilities. Firstly, it is established that adult second language learners struggle to attain native-like proficiency in gender assignment \citep{Unsworth2008AgeDutch,Bartning2000GenderLearners}. The poor performance we observe could be because children are better learners of grammatical gender than adults \citep{Blom2006EffectsInflection}. Secondly, it is established that children rely on morphological cues in noun endings for gender acquisition, while semantic cues play a more minor role \citep{Karmiloff-Smith1981AReference}. In our experiment, despite novel nouns having typically neutral endings, participants may still assign noun gender based on their intuitive familiarity with gender-typical endings, rather than adhering to the gender in the learning example. Lastly, the feminine inflections, especially in adjectives and past-participles, only result in subtle changes in pronunciation, reinforcing the tendency to default to the masculine gender category. 

\section{Conclusion}
Characterising the ability of models to generalise linguistic knowledge in a human-like way remains a challenge, and the potential impacts are twofold. In terms of the mechanistic interpretability of models, such studies lead to a better understanding of how specific linguistic generalisations are achieved. Our work shows that grammatical gender information for nouns is sufficiently encoded in word embeddings and can be used to perform agreement across syntactic contexts; however, it is unclear whether gender information is primarily hosted by the embeddings, and the specific noun, or whether other mechanisms in the model are more critical. It may be that models may not employ a genuine abstraction of grammatical gender in order to generalise gender agreement tasks to new nouns, and may employ different mechanisms for each agreement context. Further work is required to understand the mechanisms underlying our behavioural result, showing successful generalisation. 

From a psycholinguistic perspective, we find some parallels between model and human biases, and learning strategies. We find asymmetric model performance across gender categories and syntactic agreement contexts, which points to a default reasoning strategy in models \citep{Jumelet2019AnalysingAssignment}. The same behavioural pattern was also found in our human word learning experiment, supporting the default reasoning hypothesis for gender acquisition in French \citep{Boloh2010GenderChildren,Vigliocco1999WhenProduction}. More broadly, examining how humans and models employ grammatical properties in novel contexts offers possible strategies and testable hypotheses for abstract linguistic representation and generalisation in both systems.

\section{Limitations}
\paragraph{Novel-noun embeddings} Our method for creating novel nouns preserves semantics and syntactic information in the embeddings, but unlike in comparative scenarios for children learning a new word, the novel nouns are devised such that they have an initial gender categorisation. We do note that few-shot learning behaviour was still successful for novel nouns with initial gender categorisations of $49-51\%$ for either gender. In future, we aim to explore other controlled methods, such as iterative null space projection \citep{Ravfogel2020NullProjection}, to remove gender information from word-embeddings before few-shot learning. 

\paragraph{Construction of test sentences} Although care was taken to construct grammatical tests and interfering material with gender-neutral words except for the target region, agreement accuracy could have be been affected by unintended gender cues. Our method of probing gender information was through the task of simple grammatical agreement. This could be extended to include other gender agreement constructions to better quantify gender information in the word-embeddings. For example, including other determiners like `un/une' and `du/de la' and other relative pronouns. Our lists of nouns, adjectives, and participles were frequency-matched across genders, and few-shot learning behaviour was consistent in all 20 novel-noun combinations - however, future work could expand this paradigm to confirm the effect with a larger set of nouns. 

\paragraph{Evaluation} We evaluated our experimental paradigm in four gender agreement contexts and two language models; our few-shot word-learning and testing paradigm can be extended to include extensive tests of grammatical gender agreement, and more complex linguistic constructions such as nested-dependencies and testing agreement across attractor nouns with contradicting number or gender \citep{Marvin2018TargetedModels}. This framework can also be used to test grammatical abstraction in multilingual LSTMs, other Transformer architectures and the transfer of grammatical representations learnt across languages \citep{Gonen2022AnalyzingModels,Mueller2020Cross-LinguisticModels} and model architectures.

\paragraph{Tokenisation} 
We used word-by-word tokenisation to prepare the data for language modelling. However, morphology is an important aspect of French and grammatical gender. In French, nouns, adjectives and verbs are often inflected based on their gender and number. Morpho-syntactic rules are one of the main linguistic aspects underpinning grammatical generalisations learnt and employed by children \cite{Berko1958TheMorphology}. While tokenising by words provides a method for investigating the generalisation of grammatical properties of words, purely based on syntactic categories and structure, morpho-syntactic inflections are fundamental rules employed by humans. Future studies could consider whether models trained on sub-word tokenisation, taking into account the role of morpho-syntactic properties of gender, also develop a similar representation of abstract grammatical gender, and exhibit the same learning patterns and biases. 

\paragraph{Beyond French}
Future research could explore how models, compared to people, learn to represent grammatical gender and agreement rules across many languages. The grammatical gender system in each language has a different number of categories and how they interact with semantic interpretation; these manifest in different agreement rules and morphological markings. Our study focused on a typical two-gender system in French. While the gender systems of Romance languages are quite similar, an immediate next step could be to compare how two-gender systems (French, Spanish, Italian) function differently to three-gender systems like German. 

The Bantu languages present a more complex gender system; they commonly have five to ten gender categories  \citep{DiGarbo2022ASystems}. These categories are not based on biological sex; some are based on semantic categories like human/non-human and animate/inanimate, but others are more abstract. 

Relatedly, Dutch presents a challenging gender system due to its inconsistent agreement markers \citep{Audring2016Gender}; its indefinite articles and numerals do not indicate gender, and there is considerable variation in gender among relative and demonstrative pronouns \citep{Cornips2008FactorsDutch}. Without consistent agreement markers, the language model is forced more towards the abstraction of gender, which is central to the noun, as memorisation based on individual lexical units would be inefficient. Can language models develop an abstract representation of grammatical gender and agreement rules in more complex gender systems like these?

\section{Ethical Considerations}
This research characterises the capabilities of language models to learn grammatical properties. While our current study does not present any direct risks or ethical concerns, we acknowledge potential influences on broader issues such as bias and fairness. Cultural biases are often amplified by large language models \citep{Vig2020InvestigatingAnalysis} in practical inference tasks like sentiment analysis and assigning gender pronouns to professions. In our study, we observe that our language models exhibit biases in learning grammatical gender categories. We demonstrate across two very different model architectures that gender information encoded in word-embeddings can be influenced through straightforward learning updates. While this changes gender-categorisation behaviour, it does not mitigate the inherent bias as evidenced by differences in learning each category. This adds to concerns raised by \citet{Gonen2019LipstickThem} that adjusting embeddings based on the gender direction alone may not be a foolproof method for de-biasing \citep{Bolukbasi2016ManEmbeddings,Zhao2020GenderTransfer}. 

\section*{Acknowledgments}
This work was supported by the Wellcome Trust (PS) [108899/B/15/Z]; Leverhulme Trust (CH) [RF-2021-533]; Institute for Cognitive Neuroscience HSE, RF government (NK) [075-15-2022-1037]. The authors would like to thank the reviewers for their helpful feedback.

\bibliography{references_all}
\bibliographystyle{acl_natbib}

\clearpage
\appendix
\section{Training dataset}
\label{sec:appendixA}
We trained the LSTM language model described in \citep{Gulordava2018ColorlessHierarchically} on French Wikipedia data \citep{Mueller2020Cross-LinguisticModels} with the objective of next-word prediction. The original corpora was obtained from Wikipedia, marked up using WikiExtractor, and tokenized word-by-word using TreeTagger with 50,000 tokens. We further cleaned the vocabulary of 50,000 most common tokens used in \citep{Mueller2020Cross-LinguisticModels} by removing capitalisation, punctuation and tokens which were repeated due to errors in accents, resulting in 42,908 tokens. The remaining tokens in the corpus were tagged as unknown with <unk> before training. Sentences with more than 5\% unknown tokens were eliminated. Sentences were shuffled and split into training, validation, and test sets using a 8:1:1 ratio.

\section{Language models}
\label{sec:appendixB}
For our LSTM model, we follow exactly the training procedure described in \citep{Mueller2020Cross-LinguisticModels}. For the transformer, we use decoder-only model with 12 layers, 12 heads, embedding and hidden size of 768, sequence length of 100, trained with a language modelling objective where the probability of a given token is estimated knowing only the preceding tokens. As with the LSTM, the transformer's input and output embedding layers were tied. A combination of hyper-parameters were explored while training the Transformer: dropout: 0, 0.1, 0.2, batch size: 32, 64, choice of optimizer: AdamW, Stochastic Gradient Descent (SGD) with momentum and learning rate schedulers with warm ups: cosine annealing, linear decay. We chose the training protocol and hyper-parameters that provided lowest test perplexities and best performance on the baseline gender-agreement task, Section \ref{fig:baseline}. While a discussion of the choice of optimizers is beyond the scope of this work, we found that training with SGD resulted in a model that generalised better for our task, despite training with AdamW resulting in similar perplexities, Table \ref{tab:model_params}. For the final models with three random initialisation seeds, we used a linear warm up epoch with and a cosine scheduling on 50 epochs with maximum learning rate 0.02 without restarts. Compute: two NVIDIA P100 GPUs were used.
Code and data availability: \url{https://github.com/prisukumaran23/lstm_learning}
\section{Novel noun combinations}
\label{sec:appendixC}
Each row shows the feminine and masculine gendered nouns, and English translations, that were combined to create 20 novel nouns. \\

\begin{tabular}{ll}
\textbf{Feminine Noun} & \textbf{Masculine Noun}\\
assiette (plate) & bol (bowl) \\
bague (ring) & bracelet (bracelet) \\
écharpe (scarf) & foulard (scarf) \\
fourchette (fork) & fouet (whisk) \\
gomme (eraser) & stylo (pen) \\
lampe (lamp) & lustre (chandelier) \\
perle (pearl) & diamant (diamond) \\
plante (plant) & arbre (tree) \\
tarte (pie) & gâteau (cake) \\
vanne (valve) & robinet (faucet) \\
tasse (cup) & bol (bowl) \\
casquette (cap) & feutre (felt) \\
cerise (cherry) & citron (lemon) \\
colle (glue) & ruban (ribbon) \\
cuillère (spoon) & couteau (knife) \\
cuisinière (stove) & réfrigérateur (refrigerator) \\
guitare (guitar) & violon (violin) \\
perruque (wig) & bonnet (cap) \\
scie (saw) & marteau (hammer) \\
tablette (tablet) & ordinateur (computer) 
\end{tabular}

\begin{table*}[ht]
\small
\centering
\begin{tabular}{cccccccccc}
\hline
model & \begin{tabular}[c]{@{}c@{}}emb/hid \\ size\end{tabular} & layers & \begin{tabular}[c]{@{}c@{}}batch\\ size\end{tabular} & dropout & \begin{tabular}[c]{@{}c@{}}learning \\ rate\end{tabular} & \begin{tabular}[c]{@{}c@{}}best \\ epoch\end{tabular} & \begin{tabular}[c]{@{}c@{}}optimizer/\\ lr scheduler\end{tabular} & \begin{tabular}[c]{@{}c@{}}test\\ ppl\end{tabular} & \begin{tabular}[c]{@{}c@{}}accuracy on \\ baseline task\end{tabular} \\ \hline
\multirow{2}{*}{LSTM} & 650 & 2 & 128 & 0.1 & 10 & 50 & SGD/linear & 41.8 & 94.2 \\
 & 650 & 2 & 128 & 0.1 & 20 & 49 & SGD/linear & 41.6 & 96.4 \\ \hline
\multirow{4}{*}{Transformer} & 768 & 12 & 64 & 0 & 0.0005 & 38 & adamw/cosine & 32.9 & 81.1 \\
 & 768 & 12 & 64 & 0.1 & 0.0005 & 41 & adamw/cosine & 31.3 & 82.5 \\
 & 768 & 12 & 64 & 0 & 0.02 & 46 & SGD/cosine & 32.2 & 90.2 \\
 & 768 & 12 & 64 & 0.1 & 0.02 & 45 & SGD/cosine & 31.5 & 91.6 \\ \hline
\end{tabular}
\caption{Top two LSTM models and transformer models trained with SGD/AdamW and their hyperparameters, perplexities and accuracy on baseline gender agreement on existing nouns.}
\label{tab:model_params}
\end{table*}

\begin{table*}[ht]
\centering
\begin{tabular}{ll}
\toprule
\textbf{Agreement type} & \textbf{Example} \\
\midrule
noun-adjective & on ne voit pas de \underline{\textbf{table}}\textsubscript{f} [en ce moment qui est] \textit{\textbf{vert}\textsubscript{m}/\textbf{verte}\textsubscript{f}} \\
{\small temporal modifier + relative clause}
& (we do not see a \underline{\textbf{table}} [at the moment that is] \textbf{green}) \\
\addlinespace
\midrule
noun-participle & \underline{\textbf{table}}\textsubscript{f} [en ce moment qui est] \textit{\textbf{brisé}\textsubscript{m}/\textbf{brisée}\textsubscript{f}} \\
{\small temporal modifier  +  relative clause}
& (we do not see a \underline{\textbf{table}} [at the moment that is] \textbf{broken}) \\
\addlinespace
\midrule
noun-relative-pronoun & je vois l' \underline{\textbf{ampoule}}\textsubscript{f} [plus ou moins marron sur]  \textit{\textbf{lequel}\textsubscript{m}/\textbf{laquelle}\textsubscript{f}} \\
{\small adjective phrase}
& (I see the [more or less brown] \underline{\textbf{bulb}} on \textbf{which}) \\
\bottomrule
\end{tabular}
\caption{Examples of agreement sentences with five gender-neutral intervening words}
\label{tab:testdist5}
\end{table*}

\section{Learning sentences}
\label{sec:appendixD}
List of learning sentences used in each condition with feminine/masculine training versions. 
\subsection{Condition A and B: Article-noun constructions}
je vois la\textsubscript{f}/le\textsubscript{m} noun $\langle$eos$\rangle{}$\\
je jette la\textsubscript{f}/le\textsubscript{m} noun $\langle$eos$\rangle{}$\\
je tiens la\textsubscript{f}/le\textsubscript{m} noun $\langle$eos$\rangle{}$\\
on admire la\textsubscript{f}/le\textsubscript{m} noun $\langle$eos$\rangle{}$\\
on jette la\textsubscript{f}/le\textsubscript{m} noun $\langle$eos$\rangle{}$\\
on voit la\textsubscript{f}/le\textsubscript{m} noun $\langle$eos$\rangle{}$\\
on observe la\textsubscript{f}/le\textsubscript{m} noun $\langle$eos$\rangle{}$\\
nous avons vu la\textsubscript{f}/le\textsubscript{m} noun $\langle$eos$\rangle{}$\\
nous observons la\textsubscript{f}/le\textsubscript{m} noun $\langle$eos$\rangle{}$\\
nous aimons la\textsubscript{f}/le\textsubscript{m} noun $\langle$eos$\rangle{}$\\
nous avons mangé la\textsubscript{f}/le\textsubscript{m} noun $\langle$eos$\rangle{}$\\
j' ai vu la\textsubscript{f}/le\textsubscript{m} noun $\langle$eos$\rangle{}$\\
j' aime la\textsubscript{f}/le\textsubscript{m} noun $\langle$eos$\rangle{}$\\
j' ai mangé la\textsubscript{f}/le\textsubscript{m} noun $\langle$eos$\rangle{}$\\
j' observe la\textsubscript{f}/le\textsubscript{m} noun $\langle$eos$\rangle{}$

\subsection{Condition C Noun-adjective constructions}
je vois l' noun brune\textsubscript{f}/brun\textsubscript{m} $\langle$eos$\rangle{}$\\
je vois l' noun élégante\textsubscript{f}/élégant\textsubscript{m} $\langle$eos$\rangle{}$\\
je vois l' noun excessive\textsubscript{f}/excessif\textsubscript{m} $\langle$eos$\rangle{}$\\
je vois l' noun blanche\textsubscript{f}/blanc\textsubscript{m} $\langle$eos$\rangle{}$\\
je vois l' noun violente\textsubscript{f}/violent\textsubscript{m} $\langle$eos$\rangle{}$\\
je vois l' noun noire\textsubscript{f}/noir\textsubscript{m} $\langle$eos$\rangle{}$\\
je vois l' noun agressive\textsubscript{f}/agressif\textsubscript{m} $\langle$eos$\rangle{}$\\
je vois l' noun brillante\textsubscript{f}/brillant\textsubscript{m} $\langle$eos$\rangle{}$\\
je vois l' noun massive\textsubscript{f}/massif\textsubscript{m} $\langle$eos$\rangle{}$\\
je vois l' noun lumineuse\textsubscript{f}/lumineux\textsubscript{m} $\langle$eos$\rangle{}$\\
je vois l' noun colorée\textsubscript{f}/coloré\textsubscript{m} $\langle$eos$\rangle{}$\\
je vois l' noun gravée\textsubscript{f}/gravé\textsubscript{m} $\langle$eos$\rangle{}$\\
je vois l' noun sérieuse\textsubscript{f}/sérieux\textsubscript{m} $\langle$eos$\rangle{}$\\
je vois l' noun lourde\textsubscript{f}/lourd\textsubscript{m} $\langle$eos$\rangle{}$\\
je vois l' noun ancienne\textsubscript{f}/ancien\textsubscript{m} $\langle$eos$\rangle{}$

\subsection{Condition D: Noun-participle constructions}
je vois l' noun détruite\textsubscript{f}/détruit\textsubscript{m} $\langle$eos$\rangle{}$\\
je vois l' noun brisée\textsubscript{f}/brisé\textsubscript{m} $\langle$eos$\rangle{}$\\
je vois l' noun fermée\textsubscript{f}/fermé\textsubscript{m} $\langle$eos$\rangle{}$\\
je vois l' noun renversée\textsubscript{f}/renversé\textsubscript{m} $\langle$eos$\rangle{}$\\
je vois l' noun allumée\textsubscript{f}/allumé\textsubscript{m} $\langle$eos$\rangle{}$\\
je vois l' noun gelée\textsubscript{f}/gelé\textsubscript{m} $\langle$eos$\rangle{}$\\
je vois l' noun rayée\textsubscript{f}/rayé\textsubscript{m} $\langle$eos$\rangle{}$\\
je vois l' noun bloquée\textsubscript{f}/bloqué\textsubscript{m} $\langle$eos$\rangle{}$\\
je vois l' noun fermée\textsubscript{f}/fermé\textsubscript{m} $\langle$eos$\rangle{}$\\
je vois l' noun lavée\textsubscript{f}/lavé\textsubscript{m} $\langle$eos$\rangle{}$\\
je vois l' noun peinte\textsubscript{f}/peint\textsubscript{m} $\langle$eos$\rangle{}$\\
je vois l' noun pressée\textsubscript{f}/pressé\textsubscript{m} $\langle$eos$\rangle{}$\\
je vois l' noun enflammée\textsubscript{f}/enflammé\textsubscript{m} $\langle$eos$\rangle{}$\\
je vois l' noun coupée\textsubscript{f}/coupé\textsubscript{m} $\langle$eos$\rangle{}$\\
je vois l' noun écrasée\textsubscript{f}/écrasé\textsubscript{m} $\langle$eos$\rangle{}$

\section{Test sentences with distrators}
\label{sec:appendixE}
Test sentences were carefully constructed to be gender neutral apart from the critical target region. We constructed 120 test sentences: 2 sentence beginnings x 4 intervening phrases x 15 adjectives/participles in Conditions \textbf{A} and \textbf{B}, and 24 sentence beginnings x 5 intervening phrases in Conditions \textbf{C} and \textbf{D}.  All our sentences test noun gender agreement with targets without any interfering attractor nouns. The intervening words between noun and target form either an object relative clause and temporal modifier or adjective phrase, all with the main noun as the object, see Table \ref{tab:testdist5} for examples. List of all test sentences can be found here: 
\url{https://github.com/prisukumaran23/lstm_learning/tree/main/testsets}

\section{Testing gender agreement for known nouns}
\label{sec:appendixF}
Prediction of short- and long-distant gender agreement with nouns that already exist in the original training corpus was tested to ensure that the model is fundamentally able to perform grammatical agreement. 20 masculine and 20 feminine nouns that appeared more than 50 times in the training corpus were used to construct tests for grammatical gender agreement. Noun-adjective and noun-participle tests similar to Condition \textbf{A} in Table \ref{tab:tests}, were constructed with sentence beginnings `je ne vois pas de...' or `on ne voit pas de...' followed by a noun, intervening phrase in square brackets which contained 0, 1, 3, or 6 gender-neutral words, and the adjective or participle. We constructed 600 sentences for each gender category and condition with 15 different target adjectives and participles (2 x 20 nouns x 15 targets). 

Similarly, test sentences for noun-relative-pronoun agreement were constructed with eight variations of sentence beginnings followed by nouns with vowel beginnings, and 1, 3, or 5 gender-neutral words. Each bar in Figure \ref{fig:baseline} shows prediction accuracy averaged across 600 test sentences (2 x 20 nouns x 15 target) for noun-adjective and noun-participle agreement and 160 test sentences (8 x 20 nouns) for noun-relative-pronoun construction. Examples of sentences with five gender-neutral intervening words are presented in Table \ref{tab:testdist5}.

\section{Few-shot results for Condition A split by short vs. long distance agreement}
\label{sec:appendixG}
For the LSTM, Conditions \textbf{B}, \textbf{C} and \textbf{D}, but not \textbf{A}, few-shot learning performance was consistent across test sentences with 0-6 intervening words between noun and agreement target. In Condition \textbf{A}, prediction accuracy drops by more than $10\%$ only for feminine learning trials, while there was no degradation in prediction accuracy for masculine learning trials. This is consistent with the performance difference between gender categories seen in the baseline gender agreement with existing nouns, as see Section~\ref{sec:knownnouns}.

\section{Details of human behavioural experiment}
\begin{figure*}[ht]
\centering
\includegraphics[width=\linewidth]{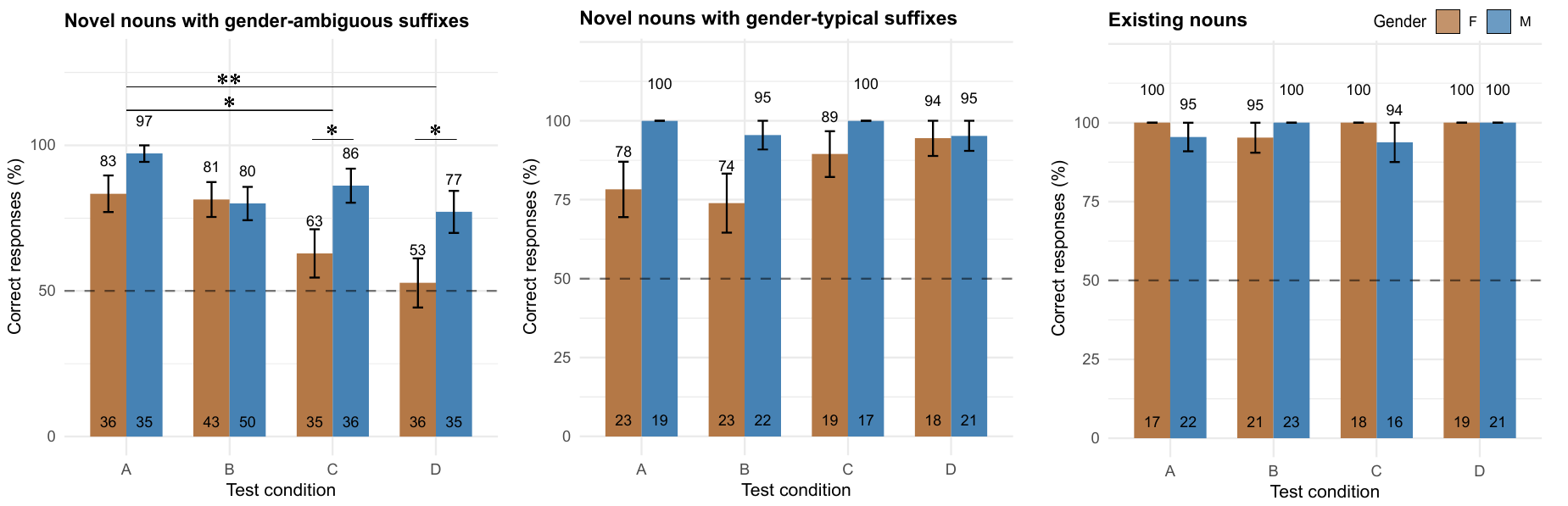}
\caption{Results of human experiment. Graphs show percentage of correct responses for gender agreement in Conditions A, B, C and D. The number of trials analysed after exclusions is shown on the bottom of each bar. \textbf{(Left)} Performance for novel nouns with ambiguous suffixes (noun endings) shows a clear masculine bias; accuracies were above 75\% in all cases except for feminine noun-relative-pronoun agreement which was near chance: $62.9\%\pm0.08$ in Condition C and ($52.8\%\pm0.08$) in Condition D. \textbf{(Middle)} Performance for novel nouns with typically feminine or masculine endings is on average ($77.6\%\pm0.05$) higher than novel nouns with ambiguous endings ($90.8\%\pm0.03$), again with higher accuracies for nouns with typically masculine endings. \textbf{(Right)} Gender agreement performance on existing nouns was very strong ($98.1\%\pm0.01$) with no marked difference between gender categories.}
\label{fig:human_exp}
\end{figure*}

We conducted an online experiment where participants learnt 16 novel nouns with gender-ambiguous endings shown in Table \ref{tab:human_exp_nouns}, adopted from \citet{Seigneuric2007TheFrench}. The nouns were presented in learning and test constructions, similar to descriptions in Table \ref{tab:tests}. During testing, they were asked to complete a test sentence with the novel noun which required grammatical gender agreement. The learning example remained visible on the screen to alleviate the load on memory, see supplementary Figure \ref{fig:human_exp_slides}. The sentence completion task was chosen to investigate intuitive responses to gender agreement. However, this meant that responses which did not match the target we were looking for, for example adjectives in Condition A and participles in Condition B, had to be excluded (see Figure \ref{fig:human_exp}). 

A total of 25  native French speakers, monolinguals, participated in the study: 9 females, 16 males, aged $M=34.4$. The experiment and participant recruitment was all conducted online on \texttt{prolific.co}. Experiments were approved by the Research Ethics Committee of the authors' main University and were performed in accordance with relevant guidelines and regulations. Participants provided informed consent prior to agreeing to take part in the online experiment, after reading instructions about the study. 

Participants underwent a total of 32 trials which were counterbalanced across conditions (A/B/C/D) and gender (F: Feminine / M: Masculine):
\begin{itemize}
    \item 16 trials for novel nouns with ambiguous gender endings, two trials for each condition (A/B/C/D) and gender (F/M) which was determined in the learning constructions
    \item 8 trials for novel nouns with typical feminine and masculine endings, one trial for each condition (A/B/C/D) and gender (F/M)
    \item 8 trials for existing nouns, one trial for each condition (A/B/C/D) and gender (F/M)
\end{itemize}
Stimuli and details of human experiment: \url{https://github.com/prisukumaran23/lstm_learning/tree/main/human_experiment}
\label{sec:appendixH}

\begin{figure*}[ht]
  \centering
  \includegraphics[width = 0.8\linewidth]{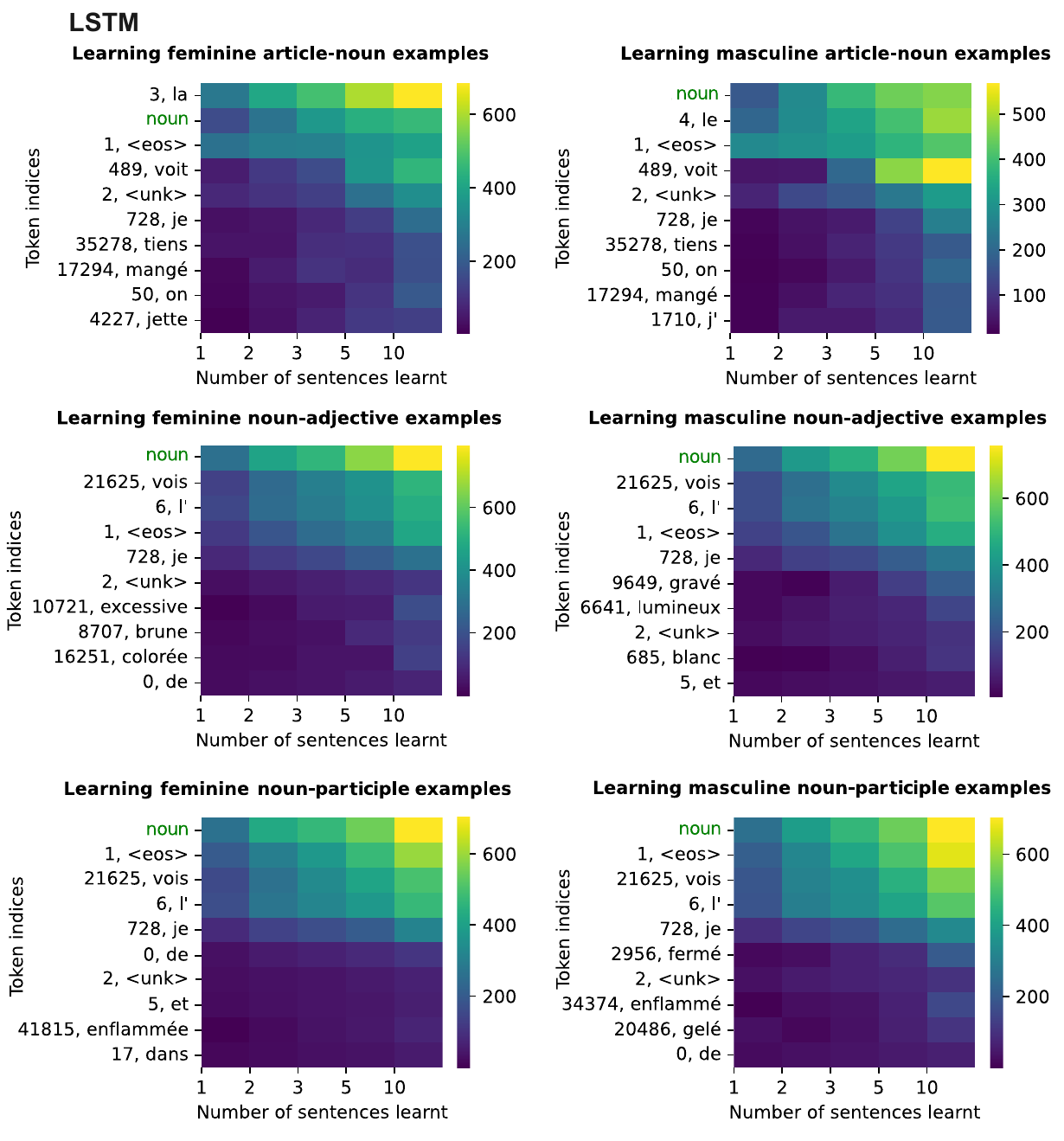}
  \caption{LSTM: Top 10 tokens by percentage of weight change to embedding layer after few-shot learning updates. Each panel shows weight changes for 1-10 learning constructions indicating feminine or masculine noun novel gender with sentence constructions from each test condition: A/B article-noun (top), C noun-adjective (mid) D noun-participle (bottom). See Table \ref{tab:tests} for learning constructions. Top tokens include the novel noun highlighted in green and other expected words from the learning examples. Note that the percentage change color scales are different in each panel.}
\label{fig:lstm_dw}
\end{figure*}

\begin{figure*}[ht]
  \centering
  \includegraphics[width = 0.8\linewidth]{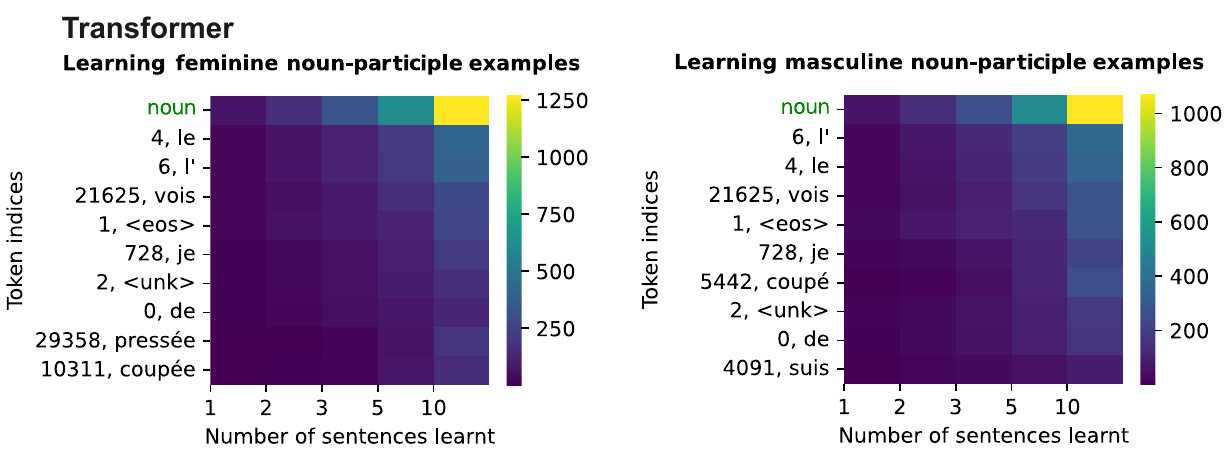}
  \caption{Transformer: Top 10 tokens by percentage of weight change to embedding layer after few-shot learning updates. Each panel shows weight changes for 1-10 learning constructions indicating feminine or masculine noun novel gender with noun-participle agreement. See Table \ref{tab:tests} for learning constructions. Top tokens include the novel noun highlighted in green. Note that the percentage change color scales are different in each panel.}
\label{fig:transformer_dw_D}
\end{figure*}

\begin{figure*}[ht]
\centering
\begin{subfigure}{7cm}
  \centering
\includegraphics[width = 6.5cm]{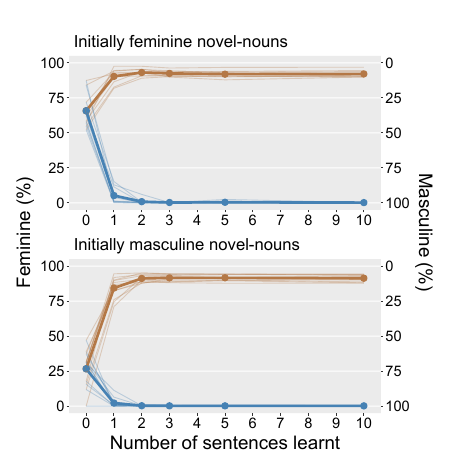}
  \caption{Agreement across 0-3 interfering words}
  \label{fig:sub1}
\end{subfigure}%
\begin{subfigure}{8cm}
  \centering
\includegraphics[width = 6.5cm]{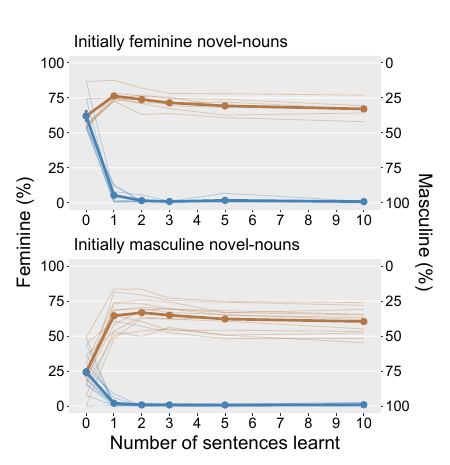}
  \caption{Agreement across 6 interfering words}
  \label{fig:sub2}
\end{subfigure}
\caption{Performance on gender agreement tests in Condition \textbf{A} with adjacent agreement \textbf{(left)} and agreement across six interfering words \textbf{(right)}. Agreement performance is shown for zero sentences learnt, and after few-shot learning with 1, 2, 3, 5 and 10 sentences. The thick orange lines indicate average prediction accuracy after learning from feminine sentences, while the blue lines correspond to learning from masculine sentences. The thin lines indicate the individual performance of 20 novel nouns. The left y-axis shows the prediction accuracy for feminine gender, while the right y-axis displays masculine gender accuracy such that 100\% accuracy for feminine gender corresponds to 0\% for masculine gender. Error bars of 95\% bootstrapped confidence intervals may be too small to be seen.}
\label{fig:longshort_agr}
\end{figure*}

\begin{table*}[ht]
\centering
\small
\begin{tabular}{@{}lll@{}}
\toprule
Ambiguous ending & Feminine/Masculine ending & Existing nouns \\ \midrule
couvirache & tamunine (F) & fleur (F) \\
spadique & viramette (F) & montagne (F) \\
sounale & l'audrelle (F) & l'étoile (F) \\
rachire & l'oivotte (F) & l'abeille (F) \\
bicatique & golcheau (M) & chien (M) \\
liavrole & forzin (M) & parapluie (M) \\
fradique & l'ousatier (M) & l'oiseau (M) \\
chonlige & l'avouguin (M) & l'ordinateur (M) \\
l'ounale &  &  \\
l'irguiste &  &  \\
l'ulole &  &  \\
l'ouchiste &  &  \\
l'aratole &  &  \\
l'aplichale &  &  \\
l'ougole &  &  \\
l'anochiste &  &  \\ 
l'anochiste &  &  \\ 
\bottomrule
\end{tabular}
\caption{List of nouns used in the human experiment, adopted from \citet{Seigneuric2007TheFrench}. Existing nouns and novel nouns with typical gender endings are labelled with F: Feminine and M: Masculine.}
\label{tab:human_exp_nouns}
\end{table*}

\begin{figure*}[ht]
\centering
\includegraphics[width=0.9\linewidth]{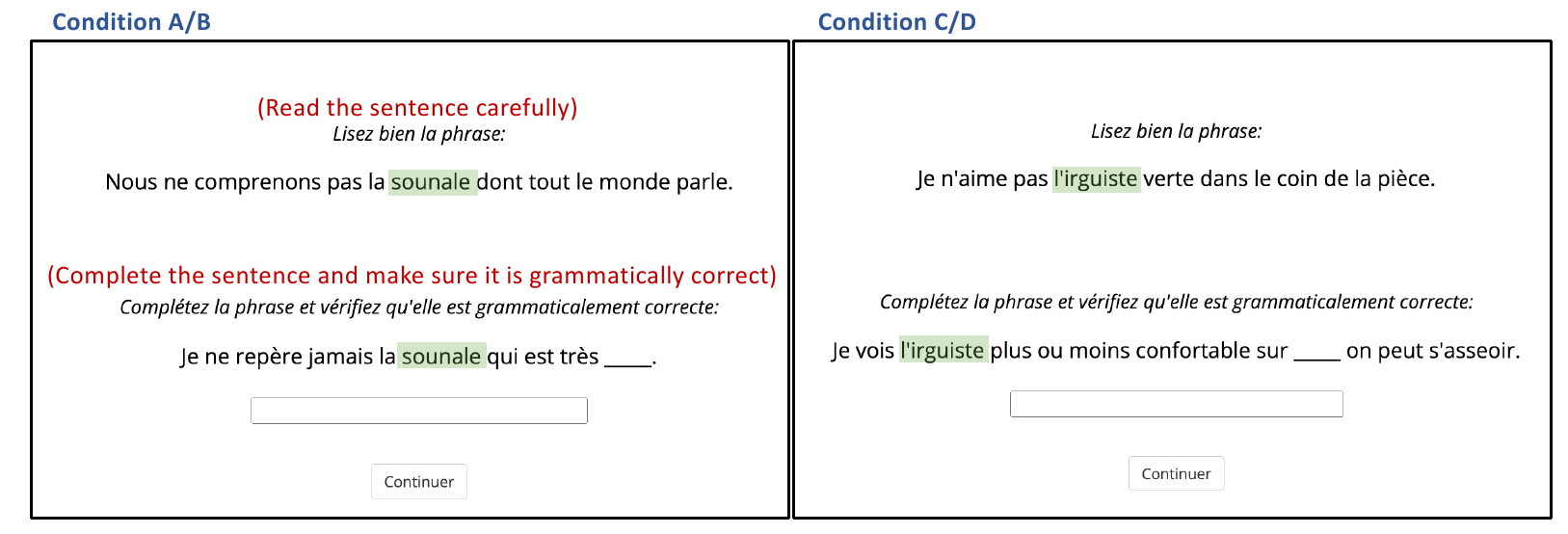}
\caption{Example screenshots of online human experiment with English translations in red text. The novel noun is highlighted in green and is an example of a noun with a gender-ambiguous ending. The same trial design was used for nouns with typical feminine or masculine gender endings, and existing nouns. The left panel shows an example of Condition A/B and right panel shows Condition C/D, analogous to those used for the language model in Table \ref{tab:tests}.}
\label{fig:human_exp_slides}
\end{figure*}

\begin{figure*}
\centering
\includegraphics[width=0.7\linewidth]{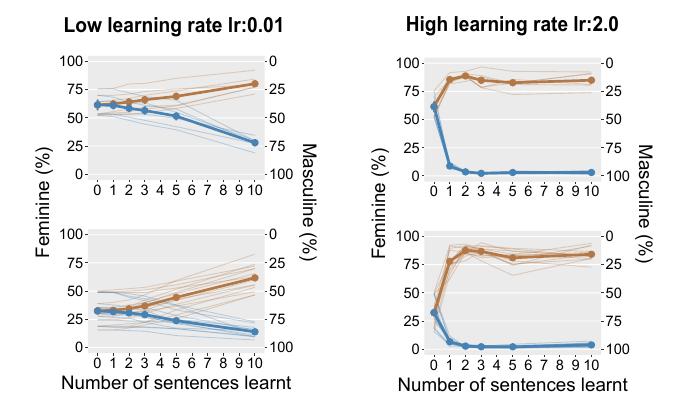}
\caption{Results of few-shot learning for the transformer language model, with low (0.01) and high (2.0) learning rates for the SGD optimizer.}
\label{fig:learningrate}
\end{figure*}

\end{document}